\def\year{2022}\relax
\documentclass[letterpaper]{article} 
\usepackage{aaai22}  
\usepackage{times}  
\usepackage{helvet}  
\usepackage{courier}  
\usepackage[hyphens]{url}  
\usepackage{graphicx} 
\usepackage{natbib}  
\usepackage{caption} 
\DeclareCaptionStyle{ruled}{labelfont=normalfont,labelsep=colon,strut=off} 
\usepackage{algorithm}
\usepackage{algorithmic}

\usepackage{bm}
\usepackage{bbm}
\usepackage{amsthm,amsmath,amssymb}
\usepackage{mathrsfs}
\usepackage{booktabs}
\usepackage{multirow}
\usepackage{bbding}
\usepackage{makecell}
\usepackage[switch]{lineno}  %
\usepackage{romannum}
\usepackage{enumitem}
\usepackage{caption}
\usepackage{subcaption}

\usepackage{newfloat}
\usepackage{listings}
\floatstyle{ruled}
\newfloat{listing}{tb}{lst}{}
\floatname{listing}{Listing}
\title{Detecting Human-Object Interactions \\ with Object-Guided Cross-Modal Calibrated Semantics}
\author{
    Hangjie Yuan,\textsuperscript{\rm 1}
    Mang Wang,\textsuperscript{\rm 3}
    Dong Ni,\textsuperscript{\rm 1,2}\thanks{Corresponding author.}
    Liangpeng Xu\textsuperscript{\rm 3}\\}
\affiliations {
    \textsuperscript{\rm 1} College of Control Science and Engineering, Zhejiang University, Hangzhou, China \\
    \textsuperscript{\rm 2} State Key Laboratory of Industrial Control Technology, Zhejiang University, Hangzhou, China \\
    \textsuperscript{\rm 3} DAMO Academy, Alibaba Group, China\\
    \{hj.yuan, dni\}@zju.edu.cn, \{wangmang.wm, liangpeng.xlp\}@alibaba-inc.com\\
}

\usepackage{bibentry}

\begin{document}
\maketitle

\begin{abstract}
    Human-Object Interaction (HOI) detection is an essential task to understand human-centric images from a fine-grained perspective. Although end-to-end HOI detection models thrive, their paradigm of parallel human/object detection and verb class prediction loses two-stage methods' merit: object-guided hierarchy. The object in one HOI triplet gives direct clues to the verb to be predicted. In this paper, we aim to boost end-to-end models with object-guided statistical priors. Specifically, We propose to utilize a Verb Semantic Model (VSM) and use semantic aggregation to profit from this object-guided hierarchy. Similarity KL (SKL) loss is proposed to optimize VSM to align with the HOI dataset's priors. To overcome the static semantic embedding problem, we propose to generate cross-modality-aware visual and semantic features by Cross-Modal Calibration (CMC). The above modules combined composes Object-guided Cross-modal Calibration Network (OCN). Experiments conducted on two popular HOI detection benchmarks demonstrate the significance of incorporating the statistical prior knowledge and produce state-of-the-art performances. More detailed analysis indicates proposed modules serve as a stronger verb predictor and a more superior method of utilizing prior knowledge. The codes are available at \url{https://github.com/JacobYuan7/OCN-HOI-Benchmark}.

\end{abstract}

\section{Introduction}
Human Object Interaction (HOI) detection has recently become a thriving research topic as it provides a fine-grained understanding to human-centric images. HOI detection aims to detect triplets formulated as $\langle$\textit{human, verbs, object}$\rangle$. Elaborate HOI detection can boost the results of image captioning \cite{yao2018exploringVR}, image retrieval \cite{johnson2015imageRt}, activity recognition \cite{yuan2021DIN}, \textit{etc.}.

Recently proposed end-to-end methods \cite{liao2020ppdm,kim2021hotr,tamura2021qpic} for HOI detection have achieved notable results without two-stage processing. Unlike two-stage methods that deal with human/object detection and verb prediction sequentially, end-to-end methods run these two processes in parallel. Since two-stage methods identify object class first, it is more effective than end-to-end methods in leveraging the object class information for verb prediction. Objects can reveal direct clues to the interactiveness of HOI triplets \cite{li2019interactiveness} and furthermore the specific interactions of HOI triplets. Although those unlikely object-verb pairs can be rejected by introducing a post-processing (\textit{e.g.} applying a binary mask \cite{tamura2021qpic}), a better choice is to inject statistical prior information during training, thus producing better ranking results \cite{chen2019KERN}. In this paper, we take a step further to implant the prior knowledge into an end-to-end model.





To profit from the object-guided hierarchy, we propose to utilize the semantic space \cite{rahman2020anyshot,xu2019HOIwithknowledge}. Semantic embeddings are intialized from word embeddings \cite{pennington2014glove,mikolov2013word2vec}. Several attempts on utilizing semantic space in HOI constrained in vanilla pretrained embeddings or their independent projections (\textit{i.e.} MLPs) \cite{zhong2020polysemy,gao2020DRG,bansal2020functionalgeneralization,peyre2019HOIwithanalogy}. However, semantic space tends to have a domain discrepancy towards visual space \cite{zhu2021semanticrelation}, which can not be overcome by such transformations. Some works \cite{xu2019HOIwithknowledge,peyre2019HOIwithanalogy} incorporate semantic space to push visual space closer to it, which will cause negative effects when visual features are strong and robust. Inspired by \cite{wu2018nonparametric_instance,you2020multilabelMSE}, we propose a Verb Semantic Model (VSM) that outputs a set of semantic features that fit in with the HOI dataset's verb co-occurrence priors. This procedure is optimized by the proposed Similarity KL (SKL) loss without entry relaxation. Then, we inject object-verb hierarchical priors via semantic aggregation, which generates a set of semantic features corresponding to a set of visual features. By proper utilization of two-modal features, performance can still be boosted under strong vision models. 

As VSM is shared across all images for a given dataset, semantic aggregation gathers static verb semantic embeddings given an object class. This static property is identical to previous methods \cite{liu2020FCMFNet} and lacks cross-modality-aware representation. As two modalities provide complementarity, we manage to solve the problem by cross-modal one-to-one calibration: mutually calibrating features from one modality via excitation from the other modality. By this mutual calibration, our model can generate a vision-aware semantic feature set and a semantic-aware visual feature set. Specifically, we propose to do Cross-Modal Calibration (CMC), which includes \textbf{i)} calibrating each modality's features by the other modality via Inter-modal Calibration (InterC), \textbf{ii)} furthermore utilizing Intra-modal Enhanced Calibration (IntraEC) \cite{AttentionAlluNeed,lin2020gps} to achieve intra-modal global reasoning. 

In order to apply the proposed VSM and CMC into practice, we select an end-to-end vision model (VM) \cite{zou2021HOITransformer,tamura2021qpic} as our VM and compose Object-guided Cross-modal Calibration Network (OCN). To conclude, our contributions are three-fold:
\begin{itemize}
    \item We introduce the object-guided statistical priors to facilitate end-to-end HOI detection. We introduce a Verb Semantic Model and use semantic aggregation to profit from this object-guided hierarchy. SKL loss is proposed to optimize VSM to align with the HOI dataset’s priors.
    
    \item To overcome the problem of static semantic embeddings, we propose to generate cross-modality-aware visual and semantic features by Cross-Modal Calibration, which consists of Inter-modal Calibration and Intra-modal Enhanced Calibration.
    
    
    \item Equipped with proposed modules, our end-to-end HOI detection model OCN achieves state-of-the-art results on two popular benchmarks. More detailed analysis indicates proposed modules serve as a stronger verb predictor and a more superior method of utilizing prior knowledge.
    
    
    
\end{itemize}

\section{Related Work}

\noindent\textbf{Human-Object Interaction Detection} Since the proposition of visual semantic role labeling and V-COCO dataset \cite{gupta2015VisualSemanticRole}, HOI detection has been a research hotspot. The methods to tackle with this problem can be categorized into two-stage methods and one-stage methods. 

Two-stage methods follow a pipeline of detecting objects first and then use cropped features to infer one HOI triplet's multi-label verb interactions. The inference usually contain multiple streams \cite{chao2018learningtodetectHOI}: an object stream, a human stream and an interaction stream, which is the key stream. InteractNet \cite{kaiming18DetectHOI} is propose to predict action-specific density maps. Contextual attention \cite{wang2019deepcontextual} is proposed to select contextual interaction information. IDN \cite{li2020hoianalysis} is proposed to learn interactions by pair integration and decomposition. Various graph-based models like GPNN \cite{Qi2018GPNN}, RPNN \cite{Zhou2019RPNN}, VSGNet \cite{ulutan2020VSGNet}, DRG \cite{gao2020DRG}, CHG \cite{wang2020CHG} are proposed to capture the interaction pattern from different aspects. Other cues like poses \cite{gupta2019nofrills,li2019interactiveness}, spatial layouts \cite{gao2018ican}, action co-occurrence \cite{kim2020actioncooccur} and language features \cite{liu2020FCMFNet,xu2019HOIwithknowledge,peyre2019HOIwithanalogy,zhong2020polysemy} are utilized to augment HOI detection.


One-stage methods can be categorized into three types: \textbf{i)} point-based methods \cite{liao2020ppdm,zhong2021GGNet} which infers heuristically-defined interaction points, \textbf{ii)} anchor-based methods \cite{kim2020uniondet} which detects union boxes, \textbf{iii)} DETR-based \cite{carion2020DETR} methods. Thanks to DETR and its ability to extract contextual cues \cite{AttentionAlluNeed}, several customized end-to-end models \cite{kim2021hotr,zou2021HOITransformer,chen2021ASNet,tamura2021qpic} originated from DETR have achieved promising results. However, there has not been an end-to-end method to incorporate semantics or explicitly utilize object-guided 
hierarchical relation priors.

\noindent\textbf{Language Semantics for Vision} The language semantics has been widely exploited in many vision subareas including zero-shot object detection \cite{bansal2018ZSOD}, zero-shot recognition \cite{wang2018ZSRsemantic}, few-shot object detection \cite{zhu2021semanticrelation} and HOI detection \cite{peyre2019HOIwithanalogy,xu2019HOIwithknowledge}. A typical method that the above works adopt is to push vision space closer to semantic space, which fails to provide positive results when the vision model is strong. In this paper, we present a method that utilizes language semantics with dataset-specific prior knowledge, which is better than the supervision from vanilla multi-modal joint embeddings \cite{xu2019HOIwithknowledge}.

\noindent\textbf{Cross-Modal Interaction} Cross-modal interaction is widely studied in cross-modal retrieval \cite{wang2019camp,liu2019referring_grounding} and VQA \cite{Gao2019DynamicInterIntra,jiang2020videoQA}. A typical practice of cross-modal interaction is to use the attention mechanism \cite{AttentionAlluNeed} to achieve cross-modal global context aggregation. While in our framework, the features in one modality will be dominated by \textit{background} class \cite{carion2020DETR}, which may degrade the quality of cross-modal context aggregation. Hence, we drop this paradigm. Instead, we introduce Cross-Modal Calibration to better incorporate cross-modal supervision for HOI. 


\section{Methodology}
In this section, we detail Object-guided Cross-modal Calibration Network (OCN) by introducing its overall pipeline, details of Verb Semantic Model (VSM) and Cross-Modal Calibration (CMC), and the training and inference strategy.

\begin{figure*}[t]
\centering
\includegraphics[width=0.8\textwidth]{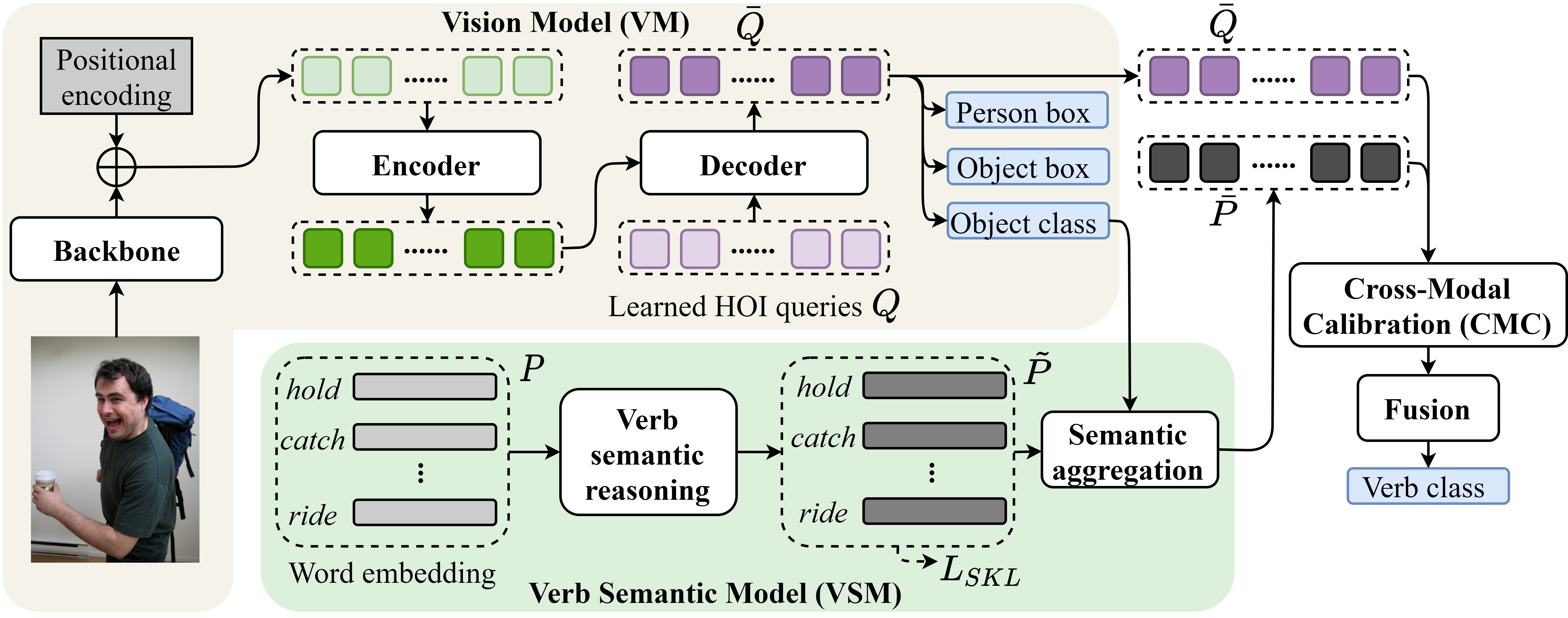} 
\caption{The overall pipeline of OCN. Generally, it consists of VM, VSM and CMC. Input word embeddings are shared across different images for a given dataset.}
\label{Overall_pipeline}
\end{figure*}

\subsection{Overall Pipeline}
The overall pipeline of OCN is shown in Figure \ref{Overall_pipeline}, which is mainly composed of three parts: a Vision Model (VM), a Verb Semantic Model (VSM) and CMC that benefits from two-modal features. For VM, we refer to the DETR-based HOI models \cite{zou2021HOITransformer,tamura2021qpic} due to their compact structure. Given an input image, VM adopts a backbone (\textit{e.g.} ResNet-50 \cite{he2016resnet}) to extract image features. We add fixed positional encoding \cite{gehring2017SeqtoSeq} to the reduced features. Then the features are flattened and fed to a Transformer encoder \cite{AttentionAlluNeed}. We define a set of learned HOI queries $\bm{Q} = \{\bm{q}_i \in \mathbb{R}^{D}\}_{i=1}^{N_q}$ to perform decoding, producing decoded query features $\bar{\bm{Q}} = \{\bar{\bm{q}}_i \in \mathbb{R}^{D}\}_{i=1}^{N_q}$, where $N_q, D$ denote the number of queries and the dimension of queries respectively. Note that one HOI query is responsible for the detection of one HOI triplet \textit{$\langle$human, verbs, object$\rangle$}. Thus, the decoded feature $\bar{\bm{q}}_i$ is then fed to independent Feed-Forward Networks (FFN) to predict a human box, an object box and an object class. 

VSM mainly consists of a verb semantic reasoning module. The inputs of VSM are word embeddings. To partly close the gap between initial word embeddings and the HOI dataset, a newly-proposed SKL loss is used to optimize the embeddings with verb co-occurrence priors. To benefit the verb prediction, we use object-guided verb semantic aggregation to generate a set of semantic embeddings $\bar{\bm{P}}$. Combined with $\bar{\bm{Q}}$ and $\bar{\bm{P}}$, we can perform CMC. After fusing features from two modalities, another FNN is applied to predict verb classes.


\subsection{Verb Semantic Model}
VSM aims to create a semantic space that is aligned with the training set. Word embeddings from pretrained models \textit{e.g.} GloVe \cite{pennington2014glove} inherently contain co-occurrence priors aligned with certain data \textit{e.g.} Google News, which may not fit in with our dataset. In VSM, we explicitly inject dataset-specific co-occurrence priors via verb semantic reasoning and SKL Loss.

To facilitate the semantic space to better close the discrepancy, we adopt a graph formulation for projection. Suppose the initialized word embeddings are $\bm{P} = \{\bm{p}_i \in \mathbb{R}^{D_p} \}_{i=1}^{N_p}$, where $N_p, D_p$ denote the number of verbs in the HOI dataset and the dimension of word embeddings. Note that word embeddings are $\ell_2$-normed. We obtain $\Tilde{\bm{P}} = \{\Tilde{\bm{p}_i} \in \mathbb{R}^{D} \}_{i=1}^{N_p}$ after verb semantic reasoning by 
\begin{equation}
    r_{ij} = \frac{\theta(\bm{p}_i)^{\rm T} \phi(\bm{p}_j)}{\sqrt{D}}; \quad r_{ij}^{'} = {\rm softmax}_{j}(r_{ij})
\end{equation}
\begin{equation}
    \Tilde{\bm{p}}_i = \sigma\left(\sum\nolimits_{j=1}^{N_p}{r_{ij}^{'} \bm{W}_{p1} \bm{p}_j}\right) + \bm{W}_{p2} \bm{p}_i
\end{equation}
where $\sigma$ is the non-linear activation function ReLU; $\theta,\phi$ are both linear projections that embed $\bm{p}_i$ into $D$-dimension space; $r_{ij}$ denotes pairwise relation; ${\rm softmax}_j$ denotes softmax function conducted along index $j$; $\bm{W}_{p1},\bm{W}_{p2} \in \mathbb{R}^{D \times D_p}$ are linear projections for value embedding and residual connection.

\subsubsection{Similarity KL} To inject priors into the semantics $\Tilde{\bm{P}}$, SKL is designed to optimize the adjacency matrix of $\Tilde{\bm{P}}$ to obey the co-occurrence distribution for a given HOI dataset. Due to the verb imbalanced problem, the naive joint distribution of verb pairs will be dominated by head classes. Inspired by \cite{you2020multilabelMSE}, we generate a symmetrized conditional distribution. We define the verb from the training set as a set $V = \{v_i\}_{i=1}^{N_p}$ and its conditional probability as $\bm{C} = \{\bm{c}_{ij} = \mathbb{P}(v_{j}|v_{i})|i,j=1,2,...N_p;i \ne j \}$. The symmetrized conditional distribution $\hat{\bm{C}}$ can be obtained by
\begin{equation}
    \hat{\bm{c}}_{ij} = \frac{\bm{c}_{ij} + \bm{c}_{ji}}{2 N_p}
\end{equation}
where $\hat{\bm{c}}_{ij}$ denotes the symmetrized probability for $\bm{c}_{ij}$. Note that the denominator $2 N_p$ is a normalizing factor to make $\hat{\bm{C}}$ sum to one. We use $\ell_2$-norm to normalize $\Tilde{\bm{p}}_i$. The adjacency matrix $\bm{A} = \{\bm{a}_{ij}|i,j=1,2,...N_p;i \ne j \}$ of the semantics $\Tilde{\bm{P}}$ can be obtained by
\begin{equation}
    \bm{a}_{ij} = \frac{{\rm exp}(\Tilde{\bm{p}}_i^{\rm T} \Tilde{\bm{p}}_j / \tau)}{\sum\nolimits_{k=1}^{N_p}{\sum\nolimits_{l=1,l\neq k}^{N_p}{{\rm exp}(\Tilde{\bm{p}}_k^{\rm T} \Tilde{\bm{p}}_l / \tau)}}}
\end{equation}
where $\tau$ is a temperature parameter that scales the softmax distribution of inner products of the normalized semantics \cite{wu2018nonparametric_instance}. Our aim is to push close distribution $\bm{A}$ to distribution $\hat{\bm{C}}$. We utilize the KL-divergence to achieve this purpose, which is formulated as 
\begin{equation}
    \mathcal{L}_{\rm SKL} = \mathbb{E}_{\hat{\bm{C}}}[{\rm log}(\hat{\bm{C}}) - {\rm log}(\bm{A})]
\end{equation}
By the utilization of $\mathcal{L}_{\rm SKL}$, the semantics are forced to obey the co-occurrence priors.

\subsubsection{Semantic Aggregation} To facilitate the verb prediction, we propose object-guided verb semantic aggregation. The basic intuition is that given an object class for one specific HOI triplet, it indicates what the verb might be or might not be based on the object-verb co-occurrence priors. To realize this, we first collect the conditional priors $\bm{S} = \{ \bm{s}_{ij} = \mathbb{P}(v_j|o_i) | i = 1,2,...,N_o+1; j = 1,2,...,N_p \}$ from the training set, where $o_i \in \bm{O} = \{o_i\}_{i=1}^{N_o+1}$, $N_o$ denotes the number of object classes and the additional 1 denotes the \textit{background} class. Note that $\bm{s}_{i j}|_{i=N_o+1}$ is manually set to be a uniform distribution, as the \textit{background} class gives no clue about the verb classes. To overcome the effect caused by false prediction of object classes and to alleviate the long-tailed distribution of verbs, we use Laplacian Smoothing \cite{zhai2004smoothingmethods} to smooth the conditional distribution $\bm{s}_{ij}$ as 
\begin{equation}  \label{smoothing_probability}
    \hat{\bm{s}}_{ij} = \frac{\bm{s}_{ij} + \beta / N_p}{\sum\nolimits_{k=1}^{N_p}{(\bm{s}_{ik} + \beta / N_p )}} = \frac{\bm{s}_{ij} + \beta / N_p}{1 + \beta} 
\end{equation}
where $\hat{\bm{s}}_{ij} $ denotes the smoothed object-verb priors for $\bm{s}_{ij}$; $\beta$ is a hyper-parameter for smoothing.

Given a decoded HOI query set $\bar{\bm{Q}}$, we predict their object classes by FFN. Combined with object-verb priors, we manage to aggregate a set of $\bar{\bm{P}} = \{ \bar{\bm{p}}_i \in \mathbb{R}^D \}_{i=1}^{N_q}$, which indicates a verb guess based on the smoothed priors. Note that $\bar{\bm{Q}}$ and $\bar{\bm{P}}$ are one-to-one match, and the cardinalities of $\Tilde{\bm{P}}$ and $\bar{\bm{P}}$ are different. The verb semantic aggregation process can be formulated as
\begin{equation}\label{semantic_aggregate}
    \bar{\bm{q}}_i^{(o)} = {\rm FFN}(\bar{\bm{q}}_i); \quad
    \bar{\bm{p}}_i =  \sum\nolimits_{j=1}^{N_p}{\hat{\bm{s}}_{\bar{\bm{q}}_i^{(o)}    j} \Tilde{\bm{p}}_j}
\end{equation}
where $\bar{\bm{q}}_i^{(o)}$ denotes the object class of query $\bar{\bm{q}}_i$ predicted using FFN. $\hat{\bm{s}}_{\bar{\bm{q}}_i^{(o)} j}$ is the smoothed $\mathbb{P}(v_j|o_{\bar{\bm{q}}_i^{(o)}})$. Eq.\ref{semantic_aggregate} utilizes the object class as a cue, to aggregate verb semantic embeddings based on the smoothed priors $\hat{\bm{s}}_{\bar{\bm{q}}_i^{(o)} j}$.

\subsection{Cross-Modal Calibration}
\begin{figure}[t]
\centering
\includegraphics[width=0.4\textwidth]{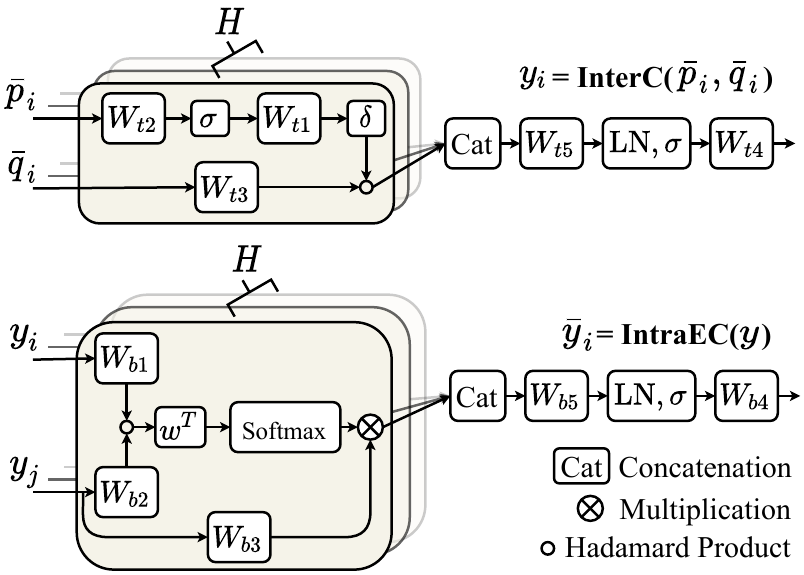} 
\caption{Example visualizations for $\bm{y}_i = {\rm InterC}(\bar{\bm{p}}_i, \bar{\bm{q}}_i)$ and $\bar{\bm{y}}_i = {\rm IntraEC}(\bm{y})$. ${\rm softmax}$ is used along index $j$.}
\label{Module_visualization}
\end{figure}

\begin{table*}[t]
    \begin{subtable}[t]{0.4\textwidth}
        \small
          \setlength{\tabcolsep}{2pt}
          \centering
            \begin{tabular}{cccc|ccc}
            \Xhline{1.0pt}
            \textbf{$\mathcal{L}_{\rm SKL}$} & \textbf{VSM} & \textbf{InterC} & \textbf{IntraEC} & \textbf{Full} & \textbf{Rare} & {\footnotesize \textbf{Non-Rare}} \\
            \hline
            \hline
            \multicolumn{4}{c|}{Base VM} & 29.15  & 22.20  & 31.23  \\
            \checkmark      & \checkmark      & \checkmark      & \checkmark      & \textbf{30.91} & \textbf{25.56} & \textbf{32.51} \\
                  & \checkmark      & \checkmark      & \checkmark      & 30.54  & 24.33  & 32.39  \\
            \checkmark      & \checkmark      &       &       & 29.88  & 23.50  & 31.79  \\
            \checkmark      & \checkmark      & \checkmark      &       & 30.51  & 25.15  & 32.12  \\
            \checkmark      & \checkmark      &       & \checkmark      & 30.40  & 24.71  & 32.10  \\
            \Xhline{1.0pt}
            \end{tabular}%
          \caption{Ablation study of proposed modules and loss.}
          \label{Ablation_module_loss}%
    \end{subtable}
    \hfill
    \begin{subtable}[t]{0.25\textwidth}
        \small
        \setlength{\tabcolsep}{2pt}
          \centering
            \begin{tabular}{c|ccc}
            \Xhline{1.0pt}
            \textbf{\#Heads $H$} & \textbf{Full} & \textbf{Rare} & {\footnotesize \textbf{Non-Rare}} \\
            \hline
            \hline
            1  & 30.58  & \textbf{25.65}  & 32.05  \\
            2  & \textbf{30.91} & 25.56 & 32.51  \\
            4  & 30.82  & 24.45  & \textbf{32.73} \\
            8  & 30.65  & 24.93  & 32.36  \\
            \Xhline{1.0pt}
            \end{tabular}%
            \caption{Effect of head numbers for both InterC and IntraEC.}
          \label{Ablation_heads}%
     \end{subtable}
     \hfill
    \begin{subtable}[t]{0.25\textwidth}
          \small
          \setlength{\tabcolsep}{2pt}
          \centering
            \begin{tabular}{c|ccc}
            \Xhline{1.0pt}
            \textbf{$\tau$}  & \textbf{Full} & \textbf{Rare} & {\footnotesize \textbf{Non-Rare}} \\
            \hline
            \hline
            0.025  & 30.70  & 25.03  & 32.39  \\
            0.050  & \textbf{30.91} & 25.56 & \textbf{32.51} \\
            0.100  & 30.54  & 24.67  & 32.29  \\
            0.200  & 30.72  & \textbf{25.59}  & 32.25  \\
            \Xhline{1.0pt}
            \end{tabular}%
          \caption{Effect of varying choices for temperature $\tau$ in $\mathcal{L}_{\rm SKL}$.}
          \label{Ablation_tau}%
     \end{subtable}
     \caption{Ablation study on HICO-DET.}
     \label{Ablation}
\end{table*}

The obtained set $\bar{\bm{Q}}$ and $\bar{\bm{P}}$ contain visual cues and semantic cues for HOI verbs respectively. To overcome the static semantic embedding problem, we propose to calibrate one modality guided by the other modality. Previous work globally aggregate features from the other modality \cite{Gao2019DynamicInterIntra} but both modality in our framework is dominated by \textit{background} class, which will not result in a fine cross-modal aggregation. Inspired by \cite{hu2018SENet,AttentionAlluNeed}, we propose to perform cross-modal feature calibration in multiple subspaces, which includes InterC and IntraEC illustrated in Figure \ref{Module_visualization}. Below we will exemplify using $\bar{\bm{p}}_i$ to calibrate $\bar{\bm{q}}_i$, denoted as $\bm{y}_i = {\rm InterC}(\bar{\bm{p}}_i, \bar{\bm{q}}_i)$ and performing intra-modal reasoning, denoted as $\bar{\bm{y}}_i = {\rm IntraEC}(\bm{y})$. The generation of $\bm{x}_i = {\rm InterC}(\bar{\bm{q}}_i, \bar{\bm{p}}_i)$ and $\bar{\bm{x}}_i = {\rm IntraEC}(\bm{x})$ is similarly conducted.

\subsubsection{Inter-Modal Calibration} To perform InterC, we project $\bar{\bm{p}}_i$ into a subspace and then excite other modality's feature $\bar{\bm{q}}_i$ in the subspace. We formulate it as 
\begin{equation}
     \bm{e}_{i} = \delta(\bm{W}_{t1} \sigma(\bm{W}_{t2} \bar{\bm{p}}_i)) \circ \bm{W}_{t3} \bar{\bm{q}}_i 
\end{equation}
where $\bm{W}_{t2},\bm{W}_{t3} \in \mathbb{R}^{\frac{D}{H} \times D}$ project $\bar{\bm{p}}_i$ and $\bar{\bm{q}}_i$ into $\frac{D}{H}$-dimension subspaces; $\bm{W}_{t1} \in \mathbb{R}^{\frac{D}{H} \times \frac{D}{H}}$ projects the $\sigma$-activated feature into $\frac{D}{H}$-dimension subspaces; $\delta$ is the sigmoid function; $\circ$ denotes Hadamard product. We project the $\bar{\bm{q}}_i$ in rather low dimension in order to perform multi-head calibration with acceptable computational cost. We can perform the calibration as
\begin{equation}
    \bm{y}_i = \bar{\bm{q}}_i + \bm{W}_{t4} \sigma({\rm LN}(\bm{W}_{t5} {\rm Cat}(\bm{e}_{i}^{(1)},...,\bm{e}_{i}^{(H)})))
\end{equation}
where $H$ denotes the number of independent InterC heads; superscript is added to $\bm{e}_{i}$, denoted as different heads; ${\rm LN}$ denotes LayerNorm \cite{ba2016layernorm}; ${\rm Cat}$ denotes concatenation; $\bm{W}_{t4},\bm{W}_{t5} \in \mathbb{R}^{D \times D}$.

\subsubsection{Intra-Modal Enhanced Calibration} Since InterC calibrates the features, it is intuitive to incorporate a global reasoning module IntraEC to restore the global context. In practice, we adopt bilinear pooling \cite{kim2016hadamard,yuan2021learningcontext} to infer intra-modal relation as
\begin{equation}
    f_{ij} = \bm{w}^{{\rm T}} (\bm{W}_{b1} \bm{y}_i) \circ (\bm{W}_{b2} \bm{y}_j)
\end{equation}
\begin{equation}
    f_{ij}^{'} = {\rm softmax}_j (f_{ij}); \quad \bm{g}_{ij} = f_{ij}^{'} \bm{W}_{b3} \bm{y}_{j}
\end{equation}
where $\bm{w} \in \mathbb{R}^{D}$, $\bm{W}_{b1},\bm{W}_{b2} \in \mathbb{R}^{D \times D}$ and $\bm{W}_{b3} \in \mathbb{R}^{\frac{D}{H} \times D}$. Also, we extend IntraEC into a multi-head manner for better representation ability:
\begin{equation}
    \bar{\bm{y}}_{i} = \bm{y}_{i} + \bm{W}_{b4} \sigma({\rm LN}(\bm{W}_{b5} \sum\nolimits_{j=1}^{N_q} {\rm Cat}(\bm{g}_{ij}^{(1)},...,\bm{g}_{ij}^{(H)})))
\end{equation}
where $\bm{W}_{b4},\bm{W}_{b5} \in \mathbb{R}^{D \times D}$; $H$ denotes the number of independent IntraEC heads, which is identical to InterC.

\subsubsection{Modality Fusion} After CMC is conducted, we obtain two sets of verified features from two modalities. We adopt the fusion strategy \cite{zhang2018learningtocount} to fuse $\bar{\bm{x}}_{i}$ and $\bar{\bm{y}}_{i}$ as  
\begin{equation}
    \bm{z}_i = \sigma(\bm{W}_x \bar{\bm{x}}_{i} + \bm{W}_y \bar{\bm{y}}_{i}) - (\bm{W}_x \bar{\bm{x}}_{i} - \bm{W}_y \bar{\bm{y}}_{i})^{2}
\end{equation}
where $\bm{W}_x,\bm{W}_y \in \mathbb{R}^{D \times D}$. The obtained feature set $\bm{Z}=\{\bm{z}_i\}_{i=1}^{N_q}$ serves as the feature for predicting verbs by FFN.

\subsection{Training and Inference Strategy} 
Similar to all DETR-based methods, we formulate HOI detection as a set prediction problem \cite{carion2020DETR}. To train the end-to-end model, we need to \textbf{i)} match ground truth (GT) HOI triplets with the predicted HOI triplets, \textbf{ii)} calculate losses for VM and VSM. The matching process is conducted with Hungarian algorithm \cite{kuhn1955hungarian}. If we denote GT HOI triplet set as $\bm{G} = \{\bm{g}_i\}_{i=1}^{N_q}$ (padded with no HOI triplets in order to match) and the predicted HOI set as $\bm{M} = \{\bm{m}_i\}_{i=1}^{N_q}$, the bipartite matching can be formulated as
\begin{equation}
    \hat{\bm{\psi}} = \mathop{\arg\min}_{\bm{\psi} \in \bm{\Psi}_{N_q}} \sum\nolimits_{i=1}^{N_q}{\mathcal{H}_{\rm cost}(\bm{g}_{i}, \bm{m}_{\psi(i)})}
\end{equation}
where $\bm{\Psi}_{N_q}$ is the solution space for bipartite matching. The matching cost $\mathcal{H}_{\rm cost}$ follows \cite{tamura2021qpic} and is detailed in the Supplementary Material. The loss to train the model can be denoted as
\begin{equation}
    \mathcal{L} = \lambda_{1} \mathcal{L}_{\rm SKL} + \lambda_{2} \mathcal{L}_{box} + \lambda_{3} \mathcal{L}_{GIoU} + \lambda_{4} \mathcal{L}_{o} + \lambda_{5} \mathcal{L}_{v}
\end{equation}
where $\mathcal{L}_{box}$ denotes $\ell_1$ loss for box regression; $\mathcal{L}_{GIoU}$ denotes GIoU loss \cite{rezatofighi2019GIoU}; $\mathcal{L}_{o}$ denotes Cross-Entropy loss for object class; $\mathcal{L}_{v}$ denotes loss for verbs. We mainly study Binary Cross-Entropy (BCE) and Focal loss \cite{lin2017focal} as $\mathcal{L}_{v}$. $\lambda$ balances these losses by setting different weights.


During inference, the object class and the bounding boxes (bbox) of the human and the object for one HOI triplet is simply generated from $\bar{\bm{q}}_i$. The object score is the maximum object confidence score. The verb score is the multiplication of the object score and the verb score predicted by $\bm{z}_i$. A binary mask is used by default to filter out object-verb pairs that are not in the training set.

\section{Experiments}
\subsection{Datasets and Metrics} 
In this paper, we use two widely-adopted datasets dubbed HICO-DET \cite{chao2015hico} and V-COCO \cite{gupta2015VisualSemanticRole}. HICO-DET contains 37,536 training images and 9,515 testing images, in which 600 $\langle$\textit{verb, object}$\rangle$ unique interaction types are defined out of 117 verb classes and 80 object classes. We evaluate on the test set by interaction mAP ($\%$) over three sets: \textbf{i)} Full set (all 600 interactions), \textbf{ii)} Rare set (138 interactions with less than 10 training samples), \textbf{iii)} Non-Rare set (462 interactions with 10 or more training samples). We evaluate our model under \textit{Default} setting for the 3 sets. One HOI triplet is rightly localized when the predicted bboxes of the human and object have Intersection-over-Union (IoU) greater than $0.5$ with GT bboxes. 

V-COCO contains 2,533 training images, 2,867 validating images and 4,946 testing images, in which interactions are defined upon 25 interactions and 80 object classes. We evaluate on the test set by verb mAP ($\%$) under two scenarios following \cite{kim2021hotr,tamura2021qpic}: \textbf{i)} In Scenario1, we need to report when there is no object in the GT HOI triplet, denoted as ${\rm AP}_{role}^{\#1}$; \textbf{ii)} In Scenario2, we can ignore the prediction of the object bbox when the GT HOI triplet is without object, denoted as ${\rm AP}_{role}^{\#2}$.

\subsection{Implementation Details} For VM, we use Transformer with a 6-layer encoder and a 6-layer decoder following \cite{carion2020DETR}. We use parameters of DETR trained on COCO \cite{lin2014MSCOCO} as VM's initialization. We adopt AdamW \cite{loshchilov2018adamW} to optimize OCN for $80$ epochs with a weight decay of $10^{-4}$. The learning rate (lr) of the backbone is fixed to $10^{-5}$. The lr of other parts starts from $10^{-4}$ and decays to $10^{-5}$ after the $60$th epoch. We use basic data augmentation to train a robust model, including random crop, random horizontal flipping, image scale and color augmentation following \cite{carion2020DETR,liao2020ppdm}. During evaluation, the most confident $K = 100$ HOI triplets are selected to compute mAP. 

By default, we use following parameters if not otherwise stated. We set number of queries $N_q=100$ and the dimension of queries $D=256$. The head number $H$ for InterC and IntraEC is set to $2$. The temperature $\tau$ of $\mathcal{L}_{\rm SKL}$ is set to $0.05$. The smoothing parameter $\beta$ is set to $0.1$. $N_p$ is the number of verb classes. The verb loss is Focal loss and the backbone is ResNet-50 by default. We use $300$-dimension GloVe word embedding \cite{pennington2014glove} as $\bm{P}$. The hyper-parameters $\lambda_1, \lambda_2, \lambda_3, \lambda_4, \lambda_5$ in $\mathcal{L}$ are set to $1,2.5,1,1,1$ respectively.


\subsection{Model Analysis}
\subsubsection{Ablation Study} We conduct model analysis on HICO-DET. Note that because VM's structure \textbf{does not change}, the ability to localize HOI triplets stays nearly unchanged. The boost purely comes from better verb-inferring ability. 

\textbf{Module and Loss Ablation} To demonstrate the efficacy of our proposed module, we first conduct ablation experiments as shown in Table \ref{Ablation_module_loss}. By applying our full model, the Rare set can gain $3.36\%$ and the Non-Rare set $1.28\%$, producing a more balanced result. If removing $\mathcal{L}_{\rm SKL}$, the model trusts the implicit co-occurrence containing in the initialized word embedding, which is not aligned with our dataset, thus causing performance decline. By applying the VSM together with $\mathcal{L}_{\rm SKL}$, the performance can be boosted by basic semantic aggregation, indicating that the object-verb prior based semantic aggregation helps to better rank the verbs. By using InterC, two modal features can calibrate features based on the other modality, bringing in performance boost. By using IntraEC, intra-modal features, especially semantic aggregated features $\bar{\bm{p}}_{i}$, can have better global-dependent representations \cite{AttentionAlluNeed}. By appending IntraEC to InterC, features in identical modality can restore the global context after calibration, bringing additive performance boost to InterC. 

\textbf{Sensitivity Analysis of Head Numbers} InterC and IntraEC are designed to be multi-head. We vary the head numbers for both modules, the results of which are shown in Table \ref{Ablation_heads}. The best choice for the head number is 2.

\textbf{Sensitivity Analysis of the Temperature Parameter} For the optimization of VSM, we vary the temperature $\tau$ in $\mathcal{L}_{\rm SKL}$ which is shown in Table \ref{Ablation_tau}. A low value of $\tau$ will sharpen the distribution of $\Tilde{\bm{A}}$ and thus ease the optimization of $\mathcal{L}_{\rm SKL}$. The table indicates the optimal value is $0.05$.

\textbf{Sensitivity Analysis of the Smoothed Distribution} An appropriate probability smoothing hyper-parameter $\beta$ in Eq.\ref{smoothing_probability} can balance the HOI detection result by smoothed semantic aggregation. We try varying $\beta$ and results are in Table \ref{smoothing_beta}. The table indicates that, \textbf{i)} compared to non-smoothed semantic aggregation, properly smoothing the object-verb distribution can slightly boost the performance; \textbf{ii)} the smoothing operation is not sensitive to the choice of $\beta$, with $\beta$ ranging from $0.1$ to $10$ producing similar results. Hence we set $\beta = 0.1$ by default; \textbf{iii)} by setting $\beta = \infty$, the model loses the verb prediction orientation, which damages the performance and proves our claim.

\begin{table}[t]
  \small
  \setlength{\tabcolsep}{5pt}
  \centering
    \begin{tabular}{c|ccc}
    \Xhline{1.0pt}
    \textbf{Smoothing $\beta$} & \textbf{Full} & \textbf{Rare} & {\footnotesize \textbf{Non-Rare}} \\
    \hline
    \hline
    0  & 30.50  & 25.00  & 32.15  \\
    0.1  & \textbf{30.91} & \textbf{25.56} & \textbf{32.51} \\
    1  & 30.85  & 25.43  & 32.47  \\
    10  & 30.81  & 25.40  & 32.43  \\
    100  & 30.36  & 24.86  & 32.01  \\
    $\infty$ & 30.18 & 23.33  & 32.22 \\
    \Xhline{1.0pt}
    \end{tabular}%
    \caption{Effect of varying $\beta$ in Eq.\ref{smoothing_probability}. $\beta = \infty$ denotes $\hat{\bm{s}}_{ij} = 1/N_p$, without any prior knowledge.}
  \label{smoothing_beta}%
\end{table}%

\begin{table}[t]
  \small
  \setlength{\tabcolsep}{3pt}
  \centering
    \begin{tabular}{cc|ccc}
    \Xhline{1.0pt}
    {\scriptsize \textbf{Verb Loss}}  & \textbf{Model} & \textbf{Full}  & \textbf{Rare}  & {\footnotesize \textbf{Non-Rare}} \\
    \hline
    \hline
    \multirow{3}[2]{*}{BCE} & VM    & $20.13$  & $13.16$  & $22.22$  \\
          & VM + MMJE & $20.60_{+0.47}$ & $13.66_{+0.50}$ & $22.67_{+0.45}$ \\
          & OCN  & $\bm{26.42}_{+6.29}$ & $\bm{20.79}_{+7.63}$ & $\bm{28.11}_ {+5.83}$ \\
    \hline
    \multirow{3}[2]{*}{Focal} & VM    & $29.15$  & $22.20$  & $31.23$  \\
          & VM + MMJE & $28.83_{-0.32}$ & $22.09_{-0.11}$ & $30.84_{-0.39}$ \\
          & OCN  & $\bm{30.91}_{+1.76}$ & $\bm{25.56}_{+3.36}$ & $\bm{32.51}_{+1.28}$ \\
    \Xhline{1.0pt}
    \end{tabular}%
    \caption{Performance analysis on HICO-DET with different verb losses and semantic models.}
  \label{Balanced_Result_and_Joint_Embedding}%
\end{table}%

\subsubsection{OCN Helps More for Poor Verb Predictor} Since our model relies on object-verb and verb-verb priors, it injects verb prediction guessing information given an object. We have reasonable speculation that adding proposed modules on a model with a poor verb predictor will have more significant improvements due to this object-guided structure. We conduct an experiment with a loss that has more trouble in inferring verbs (BCE), shown in Table \ref{Balanced_Result_and_Joint_Embedding}. BCE suffers from the problem of imbalanced positive-negative samples. OCN with BCE loss improves upon VM by $6.29\%$, suggesting that OCN greatly ameliorates the imbalance problem. Even with Focal loss, our model can also make its contribution. Moreover, the Rare set benefits more than the Non-Rare set thanks to our object-guided structure.

\subsubsection{Superiority over Multi-Modal Joint Embeddings} A common practice of utilizing word embeddings is Multi-Modal Joint Embedding (MMJE) \cite{xu2019HOIwithknowledge}, which maximizes the similarity between positive vision-semantic pairs and keep negative pairs to a predefined margin. We reimplement MMJE by VSM and $\mathcal{L}_{sim}$ in \cite{xu2019HOIwithknowledge} and add MMJE to VM, results of which are shown in Table \ref{Balanced_Result_and_Joint_Embedding}. Comparing VM and VM+MMJE, MMJE will cause negative effects when VM is already strong because it trusts the underlying verb relations in word embeddings which have a discrepancy towards the HOI dataset. However, our method with priors will constantly bring in positive effects.

\begin{table}[t]
  \small
  \setlength{\tabcolsep}{2pt}
  \centering
    \begin{tabular}{cc|cccc}
    \Xhline{1.0pt}
    \textbf{Model} & {\footnotesize \textbf{Mask}}  & \textbf{Full}  & \textbf{Rare}  & {\footnotesize \textbf{Non-Rare}} & \textbf{mR@100}\\
    \hline
    \hline
    \multirow{2}[2]{*}{VM} &       & $28.89$  & $21.76$  & $31.02$ & $57.83$ \\
          & \checkmark     & $29.15_{+0.26}$  & $22.20_{+0.44}$  & $31.23_{+0.21}$ & $65.10_{+7.27}$ \\
    \hline
    \multirow{2}[2]{*}{OCN } &       & $30.82_{+1.93}$  & $25.48_{+3.72}$  & $32.42_{+1.40}$ & $63.86_{+6.03}$ \\
          & \checkmark     & $\bm{30.91}_{+2.02}$ & $\bm{25.56}_{+3.80}$ & $\bm{32.51}_{+1.49}$ & $67.64_{+9.81}$\\
    \Xhline{1.0pt}
    \end{tabular}%
    \caption{Performance analysis on HICO-DET with different methods of utilizing prior knowledge. \textbf{Mask} denotes binary mask used to filter out impossible object-verb pairs.}
  \label{Relation_prior_mining}%
\end{table}%

\begin{table}[t]
  \small
  \setlength{\tabcolsep}{0pt}
  \centering
    \begin{tabular}{ccccc}
    \Xhline{1.0pt}
    \textbf{Method} & {\footnotesize \textbf{Backbone}} & \textbf{Time} & \textbf{\#Params} & \textbf{Full} \\
    \hline
    \hline
    PPDM \cite{liao2020ppdm}  & HOG104 & 56ms    & 194.9M  & 21.94  \\
    HOTR \cite{kim2021hotr} & R50   & 61ms    & 51.2M  & 25.10  \\
    ASNet \cite{chen2021ASNet} & R50   & 56ms    & 52.5M  & 28.87  \\
    {\scriptsize QPIC \cite{tamura2021qpic}} & R50   & 38ms    & 41.5M  & 29.07  \\
    \hline
    OCN  & R50   & 43ms    & 43.4M  & 30.91  \\
    \Xhline{1.0pt}
    \end{tabular}%
    \caption{Computational cost analysis on HICO-DET with Tesla V100. HOG and R are short for Hourglass and ResNet.}
  \label{Computation_cost}%
\end{table}%

\subsubsection{Superiority over Binary Mask} The binary mask can be treated as a stiff way to utilize relation prior knowledge by applying a hard mask to the predicted object-verb pair. Our method can be considered as a more fine-grained method which considers object-verb priors and verb co-occurrence priors. The results are shown in Table \ref{Relation_prior_mining}. In this table, we also provide results with a metric: mean Recall@$K$ (\%) \cite{tang2020unbiasedSG} (mR@$K$), which averages the recall of HOI interactions with top $K$ evaluation. $K$ is set to 100 in this paper. By applying a stiff method utilizing object-verb prior, although mR@$100$ goes up by $7.27\%$, the result of the Full set can only be boosted by $0.26\%$. While OCN w/o binary mask can boost the Full set by $1.93\%$ and the Rare set by $3.72\%$, with mR@$100$ merely going up by $6.03\%$. It indicates that OCN improves the ranking of HOI interactions greatly while the binary mask struggles to do so, demonstrating the significance of mining the object-guided verb prediction structure. If adding binary masks to OCN, the mAP improvement will be very trivial but with an obvious mR@$100$ boost, which reemphasizes the binary masks' poor ability.

\subsubsection{Computational Cost Analysis} We compare the computational cost with competitive one-stage methods, shown in Table \ref{Computation_cost}. It can be seen from the table that our method does well in speed-performance trade-off. To be more specific, VM takes 38ms to run that is identical to QPIC. VSM takes 0.6ms to run. InterC and IntraEC take 4.3ms to run. Note that during inference, we can slightly speed it up by inferring VSM once and storing $\Tilde{\bm{P}}$ in the memory for a given dataset. Our model achieves the best performance with acceptable cost adding to VM.

\begin{table}[t]
  \scriptsize
  \setlength{\tabcolsep}{1pt}
  \centering
    \begin{tabular}{ccccc|cc}
    \Xhline{1.0pt}
    \textbf{Method} & \textbf{Backbone} & \textbf{Full} & \textbf{Rare} & {\tiny \textbf{Non-Rare}} & {\tiny ${\rm AP}_{role}^{\#1}$} & {\tiny ${\rm AP}_{role}^{\#2}$} \\
    \hline
    \hline
    \multicolumn{5}{l}{\textbf{Two-stage} \quad (R = ResNet, HOG = Hourglass, ED = EfficientDet)} \\
    \hline
    InteractNet \ (\citeauthor{kaiming18DetectHOI}) & R50-FPN & 9.94  & 7.16  & 10.77 & 40.0 & - \\
    GPNN \ (\citeauthor{Qi2018GPNN}) & R152-DCN & 13.11  & 9.34  & 14.23 & 44.0 & - \\
    iCAN \ (\citeauthor{gao2018ican}) & R50 & 14.84  & 10.45  & 16.15 & 45.3 & 52.4 \\
    No-Frills\textsuperscript{*} & \multirow{2}[2]{*}{R152} & \multirow{2}[2]{*}{17.18} & \multirow{2}[2]{*}{12.17} & \multirow{2}[2]{*}{18.68} & \multirow{2}[2]{*}{-} & \multirow{2}[2]{*}{-} \\
    {\tiny (\citeauthor{gupta2019nofrills})}  &       &       &       &       &       &  \\
    PMFNet \ (\citeauthor{wan2019PMFNet}) & R50-FPN & 17.46  & 15.65  & 18.00 & 52.0 & - \\
    DRG\textsuperscript{*} \ (\citeauthor{gao2020DRG})  & R50-FPN & 19.26  & 17.74  & 19.71 & 51.0 & - \\
    IPNet \ (\citeauthor{wang2020IPNet}) & HOG104 & 19.56  & 12.79  & 21.58 & 51.0 & - \\
    FCMNet\textsuperscript{*} & \multirow{2}[2]{*}{R50} & \multirow{2}[2]{*}{20.41} & \multirow{2}[2]{*}{17.34} & \multirow{2}[2]{*}{21.56} & \multirow{2}[2]{*}{53.1} & \multirow{2}[2]{*}{-} \\
    {\tiny (\citeauthor{gupta2019nofrills})}  &       &       &       &       &       &  \\
    PD-Net\textsuperscript{*} \ (\citeauthor{zhong2020polysemy})  & R152-FPN & 20.81  & 15.90  & 22.28 & 52.6 & - \\
    IDN \ (\citeauthor{li2020hoianalysis})  & R50 & 23.36  & 22.47  & 23.63 & 53.3 & 60.3 \\
    \hline
    \multicolumn{5}{l}{\textbf{One-stage}} \\
    \hline
    UnionDet \ (\citeauthor{kim2020uniondet}) & R50-FPN & 17.58  & 11.72  & 19.33 & 47.5 & 56.2  \\
    PPDM \ (\citeauthor{liao2020ppdm}) & HOG104 & 21.94  & 13.97  & 24.32 & - & - \\
    GGNet \ (\citeauthor{zhong2021GGNet}) & HOG104 & 23.47  & 16.48  & 25.60 & 54.7 & - \\
    DIRV \ (\citeauthor{fang2020DIRV}) &  ED-d3 & 21.78  & 16.38  & 23.39 & 56.1 & - \\
    HOTR \ (\citeauthor{kim2021hotr}) & R50 & 25.10  & 17.34  & 27.42 & 55.2 & 64.4 \\
    HOITransformer \ (\citeauthor{zou2021HOITransformer}) & R50 & 23.46  & 16.91  & 25.41 & 52.9 & - \\
    QPIC\textsuperscript{\dag} & \multirow{2}[2]{*}{R50} & \multirow{2}[2]{*}{29.07} & \multirow{2}[2]{*}{21.85} & \multirow{2}[2]{*}{31.23} & \multirow{2}[2]{*}{61.8} & \multirow{2}[2]{*}{64.1} \\
    {\tiny (\citeauthor{tamura2021qpic})}  &       &       &       &       &       &  \\
    ASNet \ (\citeauthor{chen2021ASNet}) & R50 & 28.87  & 24.25  & 30.25 & 53.9 & - \\
    \hline
    \multirow{2}[2]{*}{\textbf{Ours-OCN}\textsuperscript{*}} & R50 & \textbf{30.91} & \textbf{25.56} & \textbf{32.51} & \textbf{64.2} & \textbf{66.3} \\
          & R101 & \textbf{31.43} & \textbf{25.80} & \textbf{33.11} & \textbf{65.3} & \textbf{67.1} \\
    \Xhline{1.0pt}
    \end{tabular}%
    \caption{Comparisons with state-of-the-arts on HICO-DET and V-COCO. Full, Rare and Non-Rare columns are reported on HICO-DET and ${\rm AP}_{role}^{\#1}$  ${\rm AP}_{role}^{\#2}$ on V-COCO. \textsuperscript{*} denotes utilization of word embeddings. \textsuperscript{\dag} denotes reproduction using COCO pre-trained parameters.}
  \label{SOTA_HICO_VCOCO}%
\end{table}%

\subsubsection{Object-Conditioned Verb Distribution Analysis} 
\begin{figure}[t]
\centering
\includegraphics[width=0.46\textwidth]{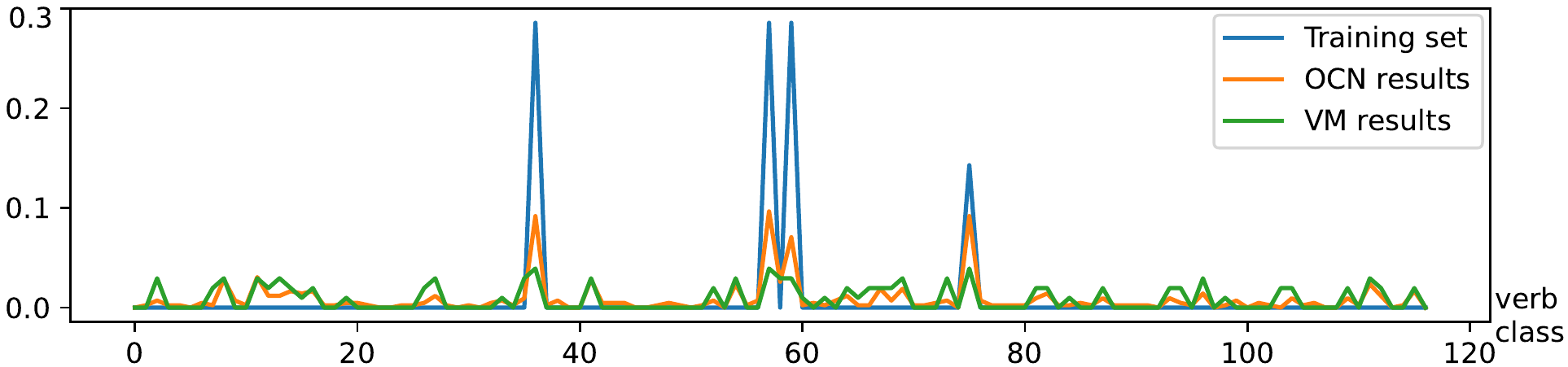} 
\caption{A case visualization of \textit{toaster}-conditioned verb distribution with different methods. Better view in color.}
\label{verb_distribution}
\end{figure}

To analyze how the object-guided hierarchy helps the verb prediction, we conduct a case study with different methods' verb distribution on the rightly localized HOI triplets whose objects are predicted as \textit{toaster}, the figure of which is illustrated in Figure \ref{verb_distribution}. Note that VM and OCN have the same vision structure, thus having identical detection abilities. VM predicts the verb more uniformly. OCN, equipped with object-guided priors, predicts the verb distribution aligned more with the distribution of the training set. We also measure the mean Pearson Correlation Coefficient (mPCC) \cite{benesty2009pearsoncoef} of the predicted object-conditioned verb distribution to the training object-conditioned verb distribution ('mean' averages different objects). It turns out that OCN has a higher mPCC (0.636) than VM does (0.476), indicating the object-guided priors help the verb prediction.

\subsubsection{t-SNE Visualization of the Semantic Space} 
\begin{figure}[t]
\centering
\includegraphics[width=0.4\textwidth]{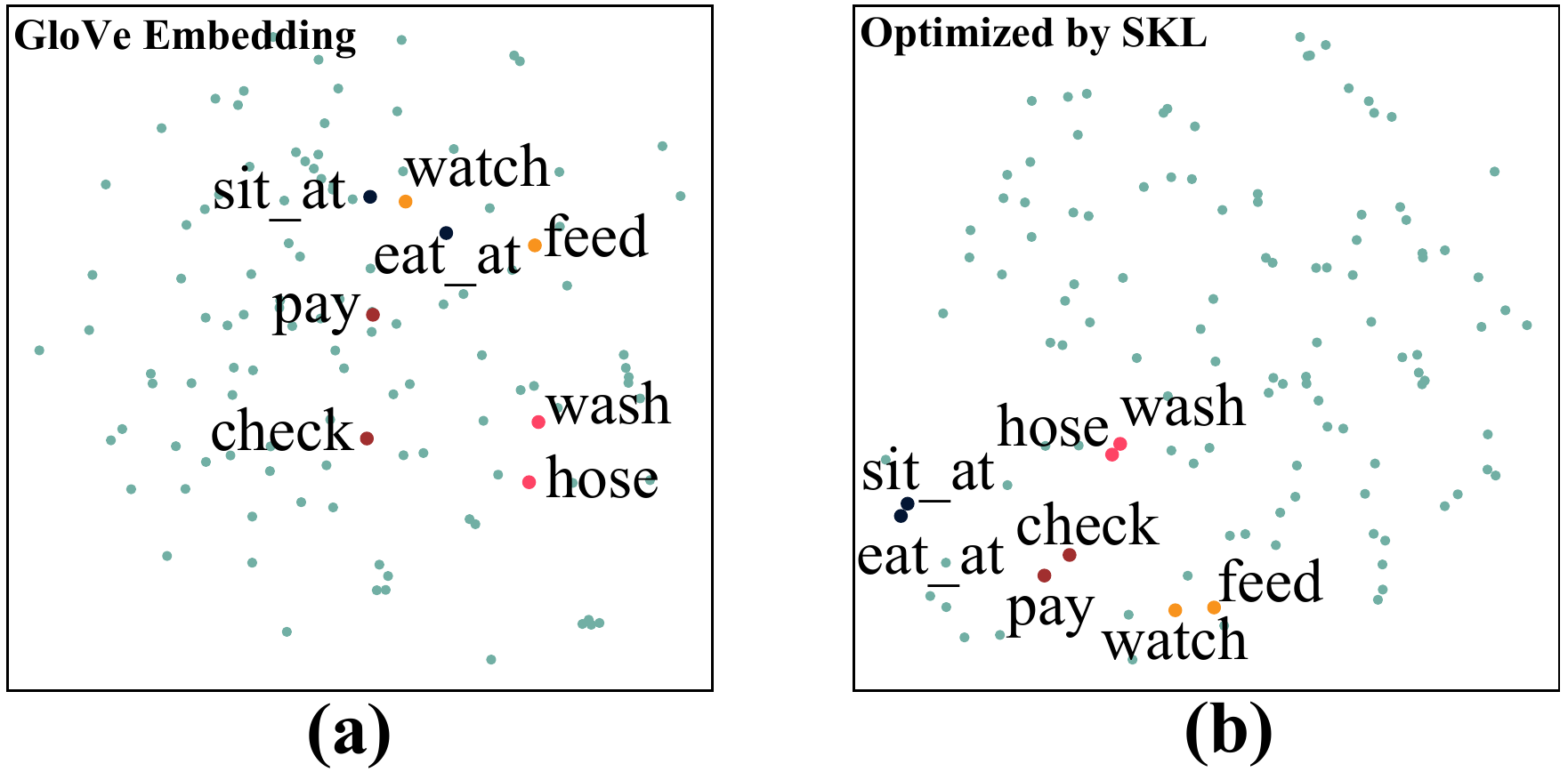} 
\caption{Visualization of the semantic space on HICO-DET.}
\label{verb_embedding}
\end{figure}

To observe SKL loss's role, we use t-SNE \cite{maaten2008visualizingt-SNE} to visualize the input Glove Embedding set $\bm{P}$ and the optimized set $\Tilde{\bm{P}}$, illustrated in Figure \ref{verb_embedding}. In Figure \ref{verb_embedding}, we highlight four word pairs with different colors that highly co-occur in HICO-DET. The original GloVe embedding spreads in the space while the optimization of SKL loss pushes semantically similar verbs closer, which decreases the discrepancy between the implicit priors in GloVe and the verb co-occurrence prior in HICO-DET.

\subsection{Comparisons with State-of-the-Arts} We compare our results with previous state-of-the-arts on HICO-DET in Table \ref{SOTA_HICO_VCOCO}. The table indicates that \textbf{i)} All previous models trail our model by a considerable margin. None of the one-stage models possesses the object-guided structure, which shows its superiority. \textbf{ii)} Comparing with previous methods that utilize word embeddings, our method greatly surpasses them by a considerable margin and maintains an end-to-end manner. \textbf{iii)} Comparing to other DETR-based models \textit{e.g.} HOTR, HOITransformer and QPIC, our method is a more balanced detector thanks to the smoothed prior knowledge.

We compare our results with previous state-of-the-arts on V-COCO in Table \ref{SOTA_HICO_VCOCO}. The table indicates our method's more superior performances compared to methods w/ or w/o word embeddings. Our method surpasses previous best ${\rm AP}_{role}^{\#1}$ by $2.4\%$ mAP and best ${\rm AP}_{role}^{\#2}$ by $1.9\%$. 


\section{Conclusions and Future Work}
In this paper, we propose to facilitate end-to-end HOI detection models with object-guided priors. In practice, we resort to a more aligned semantic space and propose to perform cross-modal calibration. Similar thoughts can also facilitate other visual relation detection problems like scene graph generation and video HOI. However, our method can run into trouble when extending to scenarios like zero-shot HOI detection, which is left for future exploration.

\section*{Acknowledgements}
We would like to appreciate Rong Jin for carefully revising the manuscript, Mingqian Tang and Jianwen Jiang for discussing the ideas, and anonymous reviewers for their valuable feedback. This work was supported by National Natural Science Foundation of China Grant No. 62173298.

\small
\bibliography{aaai22} 
\end{document}